\documentclass[5p,authoryear,preprint,12pt,twocolumn]{elsarticle}
\makeatletter\if@twocolumn\PassOptionsToPackage{switch}{lineno}\else\fi\makeatother

\usepackage{tabulary,xcolor}
\usepackage{amsfonts,amsmath,amssymb}
\usepackage[T1]{fontenc}
\makeatletter
\let\save@ps@pprintTitle\ps@pprintTitle
\def\ps@pprintTitle{\save@ps@pprintTitle\gdef\@oddfoot{\footnotesize\itshape \null\hfill\today}}
\def\hlinewd#1{%
  \noalign{\ifnum0=`}\fi\hrule \@height #1%
  \futurelet\reserved@a\@xhline}
\def\tbltoprule{\hlinewd{.8pt}\\[-12pt]}
\def\tblbottomrule{\noalign{\vspace*{6pt}}\hline\noalign{\vspace*{2pt}}}
\def\tblmidrule{\noalign{\vspace*{6pt}}\hline\noalign{\vspace*{2pt}}}

\AtBeginDocument{\ifNAT@numbers \biboptions{sort&compress}\fi}

\makeatother

\usepackage{ifluatex}
\ifluatex
\usepackage{fontspec}
\defaultfontfeatures{Ligatures=TeX}
\usepackage[]{unicode-math}
\unimathsetup{math-style=TeX}
\else 
\usepackage[utf8]{inputenc}
\fi 
\ifluatex\else\usepackage{stmaryrd}\fi

\usepackage{url,multirow,morefloats,floatflt,cancel,tfrupee}
\makeatletter

\AtBeginDocument{\@ifpackageloaded{textcomp}{}{\usepackage{textcomp}}}
\makeatother
\usepackage{colortbl}
\usepackage{xcolor}
\usepackage{pifont}
\usepackage[nointegrals]{wasysym}
\makeatletter

\def\mcWidth#1{\csname TY@F#1\endcsname+\tabcolsep}

\def\cAlignHack{\rightskip\@flushglue\leftskip\@flushglue\parindent\z@\parfillskip\z@skip}
\def\rAlignHack{\rightskip\z@skip\leftskip\@flushglue \parindent\z@\parfillskip\z@skip}

\@ifundefined{etal}{}{}

\usepackage{ifxetex}
\ifxetex\else\if@twocolumn\@ifpackageloaded{stfloats}{}{\usepackage{dblfloatfix}}\fi\fi

\AtBeginDocument{
\expandafter\ifx\csname eqalign\endcsname\relax
\def\eqalign#1{\null\vcenter{\def\\{\cr}\openup\jot\m@th
  \ialign{\strut$\displaystyle{##}$\hfil&$\displaystyle{{}##}$\hfil
      \crcr#1\crcr}}\,}
\fi
}

\AtBeginDocument{%
  \@ifpackageloaded{endfloat}%
   {\renewcommand\efloat@iwrite[1]{\immediate\expandafter\protected@write\csname efloat@post#1\endcsname{}}}{\newif\ifefloat@tables}%
}%

\def\BreakURLText#1{\@tfor\brk@tempa:=#1\do{\brk@tempa\hskip0pt}}
\let\lt=<
\let\gt=>
\def\processVert{\ifmmode|\else\textbar\fi}

\@ifundefined{subparagraph}{
\def\subparagraph{\@startsection{paragraph}{5}{2\parindent}{0ex plus 0.1ex minus 0.1ex}%
{0ex}{\normalfont\small\itshape}}%
}{}

\newcommand\role[1]{\unskip}
\newcommand\aucollab[1]{\unskip}
  
\@ifundefined{tsGraphicsScaleX}{\gdef\tsGraphicsScaleX{1}}{}
\@ifundefined{tsGraphicsScaleY}{\gdef\tsGraphicsScaleY{.9}}{}
\def\checkGraphicsWidth{\ifdim\Gin@nat@width>\linewidth
	\tsGraphicsScaleX\linewidth\else\Gin@nat@width\fi}

\def\checkGraphicsHeight{\ifdim\Gin@nat@height>.9\textheight
	\tsGraphicsScaleY\textheight\else\Gin@nat@height\fi}

\def\fixFloatSize#1{}
\let\ts@includegraphics\includegraphics

\def\inlinegraphic[#1]#2{{\edef\@tempa{#1}\edef\baseline@shift{\ifx\@tempa\@empty0\else#1\fi}\edef\tempZ{\the\numexpr(\numexpr(\baseline@shift*\f@size/100))}\protect\raisebox{\tempZ pt}{\ts@includegraphics{#2}}}}

\AtBeginDocument{\def\includegraphics{\@ifnextchar[{\ts@includegraphics}{\ts@includegraphics[width=\checkGraphicsWidth,height=\checkGraphicsHeight,keepaspectratio]}}}

\DeclareMathAlphabet{\mathpzc}{OT1}{pzc}{m}{it}

\def\URL#1#2{\@ifundefined{href}{#2}{\href{#1}{#2}}}

\def\UrlOrds{\do\*\do\-\do\~\do\'\do\"\do\-}%
\g@addto@macro{\UrlBreaks}{\UrlOrds}

\edef\fntEncoding{\f@encoding}

\makeatother

\newif\ifmultipleabstract\multipleabstractfalse%
    \makeatletter
\def\ead{\@ifnextchar[{\@uad}{\@ead}}
\gdef\@ead#1{\bgroup
   \def\_{\string\underscorechar\space}
   \def\{{\string\lbracechar\space}
   \def\textdagger{\string\textdagger\space}
   \def\texttildeapprox{\string\texttildeapprox\space}
   \def~{\hashchar\space}
   \def\}{\string\rbracechar\space}
   \edef\tmp{\the\@eadauthor}
   \immediate\write\@auxout{\string\emailauthor
     {#1}{\expandafter\strip@prefix\meaning\tmp}}
  \egroup
}
\gdef\emailauthor#1#2{\stepcounter{ead}
      \g@addto@macro\@elseads{\raggedright
      \let\corref\@gobble
      \eadsep\texttt{#1} (#2)
      \def\eadsep{\unskip,\space}}
}

\makeatother
  
\begin{document}

\begin{frontmatter}

    \title{
  \textbf{Large Language Models for Judicial Entity Extraction: A Comparative Study }    
}
    
\author[aafcb246723ab]{Atin S.Hussain\corref{c-782e2265c140}}
\ead{atin.s@u.nus.edu}\cortext[c-782e2265c140]{Corresponding author.}
\author[a588bd0c61782]{Anu Thomas}
\ead{anu\_t@sgcaruvithura.ac.in}
    
\address[aafcb246723ab]{
    National University of Singapore}
  	
\address[a588bd0c61782]{Department of Computer Applications\unskip, 
    St.George's College, Aruvithura}

\begin{abstract}
 Domain-specific Entity Recognition holds significant importance in legal contexts, serving as a fundamental task that supports various applications such as question-answering systems, text summarization, machine translation, sentiment analysis, and information retrieval specifically within case law documents. Recent advancements have highlighted the efficacy of Large Language Models in natural language processing tasks, demonstrating their capability to accurately detect and classify domain-specific facts (entities) from specialized texts like clinical and financial documents. This research investigates the application of Large Language Models in identifying domain specific entities (e.g., courts, petitioner, judge, lawyer, respondents, FIR nos.) within case law documents, with a specific focus on their aptitude for handling domain-specific language complexity and contextual variations. The study evaluates the performance of state-of-the-art Large Language Model architectures, including Large Language Model Meta AI 3, Mistral, and Gemma, in the context of extracting judicial facts tailored to Indian judicial texts. Mistral and Gemma emerged as the top-performing models, showcasing balanced precision and recall crucial for accurate entity identification.  These findings confirm the value of Large Language Models in judicial documents and demonstrate how they can facilitate and quicken scientific research by producing precise, organised data outputs that are appropriate for in-depth examination.
\end{abstract}
      \begin{keyword}
    Large language Models\sep Natural Language Processing\sep Judicial Domain\sep Judicial Entity Recognition\sep Information Extraction\sep Court Judgments
      \end{keyword}
    
  \end{frontmatter}

\section{Introduction}
Domain-specific entity recognition is a pivotal component in the realm of natural language processing, especially within specialized domains such as the legal field.  The task involves identifying and classifying judicial entities such as petitioner, respondents, and judges, attorneys etc. which are foundational for a variety of applications. These applications include relation extraction, , machine translation, sentiment analysis at the entity level, faceted search, knowledge base construction, and information retrieval\unskip~\cite{2367571:31134531} .  Finding domain-specific entities and their relationships helps improve the indexing and retrieval of legal texts and is helpful as a first step in feature selection for text clustering, classification, as well as information selection for text summarization. Furthermore, a well-tuned entity recognition(ER) system forms a basis for various applications in the legal domain as follows.

Legal Question-answering system: Judicial facts are essential in determining the responses to factoid queries. For instance, if the query is, "Who is the appellant in a particular judgement?" The answer will be predicted by the question processing module to be some judicial entity. In the event that "Mr. X" is the response and the data set has judicial entities assigned to it, the question-answering system would recognise that "Mr. X" is an entity and that it may be the response.

Creation of a knowledge graph: We can present the textual data in graphical form, such as entity-relationship graphs, if we could identify the NEs in the judicial text and the relationships among those entities.  \mbox{}\protect\newline These  graphs can be used to answer complicated relationship inquiries. Moreover, text summary is facilitated by the detection and annotation of the most pertinent information associated with a NE. \unskip~\cite{2367571:31134537}

Case-Based Reasoning: The foundation of case-based reasoning is the knowledge input that may be obtained from extracted information found in court language. This information can be fed into a variety of expert systems, including business intelligence tools and predictive analytics software.\unskip~\cite{2367571:31134625}

Relation Extraction (RE): Entity Recognition plays a pivotal role in relation extraction from judicial text by identifying key entities such as names of judges, plaintiffs, defendants, legal entities, and locations mentioned within the text. Once these entities are identified, RE identifies the relationships between them, aiding in the extraction of pertinent legal relations, such as "defendant accused of crime," "plaintiff filed a lawsuit against defendant," or "court ruled in favor of plaintiff."  \unskip~\cite{2367571:31134536}. Moreover, relation triplets can be utilized as features for other machine learning applications, such text categorization, document summarization, paraphrase detection, and so on.

The capabilities of ER systems have significantly advanced with the introduction of Large Language Models (LLMs). Equipped with advanced natural language processing methods, LLMs have shown to be extraordinarily adept in identifying and classifying objects in a wide range of complex texts. They excel at understanding and processing natural language, which makes them well-suited to handle the complexities of legal documents, which frequently include complex context and specialized~terminologies.

Through this exploration, we aim to shed light on the potential of LLMs to revolutionize ER in legal texts with zero-shot learning, paving the way for more efficient and accurate information retrieval and management within the judicial system.

The key contribution of this paper is:

\begin{itemize}
  \item \relax Evaluating the effectiveness of cutting-edge LLMs (like LLaMA 3 (Large Language Model Meta AI 3), Mistral, and Gemma) for domain-specific ER tasks within Indian legal texts.
\end{itemize}
  The paper is structured as follows. Section 2 explores the related works. Section 3 explains the state-of-the-art LLMs. Section 4 discusses the proposed methodology followed by results and discussions in section 5. Section 6 concludes the paper.

\section{Related Works}
The field of generic Named Entity Recognition (NER) has seen substantial advancements, particularly with the integration of machine learning and deep learning techniques. Early approaches relied on rule-based and statistical methods, such as Hidden Markov Models (HMMs) and Conditional Random Fields (CRFs), which, while effective to some extent, often struggled with domain-specific language and lacked generalization capabilities.

The introduction of neural network-based models marked a significant leap in NER performance. Recurrent Neural Networks (RNNs), and more specifically Long Short-Term Memory networks (LSTMs), improved the ability to capture sequential dependencies in text. The advent of attention mechanisms and Transformers further revolutionized the field, leading to the development of pre-trained language models such as BERT (Bidirectional Encoder Representations from Transformers). BERT's contextual understanding and fine-tuning abilities demonstrated remarkable improvements in NER tasks across various domains.

In the legal domain, specialized ER systems have been developed to address the unique challenges posed by legal texts, including the use of complex terminology and context-specific references. Models like Legal-BERT [\cite{2367571:31103100}] and CaseLaw-BERT [\cite{2367571:31103112}], which are pre-trained on legal corpora, have shown promise in enhancing entity recognition within legal documents. However, these models often require extensive domain-specific training data to achieve optimal performance.

Recent advancements in LLMs have further pushed the boundaries of NER capabilities. Models such as GPT-3 and its successors have exhibited exceptional proficiency in understanding and generating human-like text, which translates into improved accuracy in entity recognition tasks. The emergence of models such as LLaMA 3 and Gemma represents the latest frontier in this evolution, promising even greater performance through enhanced architectural innovations and larger training datasets. LLMs hold the added advantage of not having to be trained on legal datasets for domain-specific ER tasks.

This study builds on these advancements by evaluating the effectiveness of LLMs including LLaMA 3 [\cite{2367571:31103653}], Gemma [\cite{2367571:31104389}], Phi3 [\cite{2367571:31104390}] and Mistral [\cite{2367571:31104403}] in performing domain specific ER tasks within the context of Indian judicial texts with system prompting. By focusing on these state-of-the-art models, we aim to contribute to the growing body of research that seeks to harness the power of LLMs for specialized applications in the legal domain.

\section{Large Language Models}
This paper compares the following 4 different state-of-the-art Large Language Models in the task of domain specific  Entity Recognition for legal documents:

\begin{itemize}
  \item \relax \textbf{LLaMA 3} [\cite{2367571:31103653}] : The latest generation of Meta's open-source large language model, represents a significant advancement in natural language processing capabilities, making it highly suitable for complex tasks such as  Entity Recognition (ER) in legal documents. Featuring models with up to 70 billion parameters, LLaMA 3 excels in understanding and generating human-like text, demonstrating state-of-the-art performance across various benchmarks. Its enhanced architecture, including a more efficient tokenizer and grouped query attention, ensures superior inference efficiency and accuracy. These improvements make LLaMA 3 particularly effective in handling the specialized terminology and nuanced context typical of legal texts, thereby facilitating precise entity identification and categorization critical for legal information retrieval and document management.
  \item \relax \textbf{Gemma} [\cite{2367571:31104389}] : Developed by Google DeepMind and other teams across Google, represents a family of lightweight, state-of-the-art open models designed for high performance and broad accessibility. Available in two sizes, Gemma 2B and Gemma 7B, these models are optimized for diverse AI applications, including  Entity Recognition (ER) in legal documents. Gemma models are pre-trained and instruction-tuned, allowing them to efficiently handle the complex and domain-specific language found in case law texts. They surpass significantly larger models on key benchmarks, making them suitable for deployment on various platforms, from laptops to cloud infrastructures like Google Cloud. The incorporation of advanced fine-tuning techniques and robust evaluation processes ensures Gemma models produce safe and reliable outputs, crucial for maintaining the integrity of legal document processing. By leveraging these capabilities, the Gemma model holds promise for enhancing the accuracy and efficiency of ER tasks in the legal domain.
  \item \relax \textbf{Phi 3 }[\cite{2367571:31104390}] : The model, developed by Microsoft, represents a significant advancement in small language models (SLMs), offering exceptional performance and cost-effectiveness. Particularly relevant to  Entity Recognition  tasks in legal documents, Phi-3 models, such as the Phi-3-mini, excel due to their ability to handle long context windows up to 128K tokens. This capacity is crucial for processing extensive legal texts, ensuring comprehensive entity recognition across large document spans. Phi-3's instruction-tuned design and optimized performance across various hardware platforms, including on-device use, facilitate efficient and accurate ER in resource-constrained environments. The model's strong reasoning and logic capabilities further enhance its suitability for the analytical demands of legal document processing, providing a powerful tool for improving the efficiency and accuracy of legal information retrieval.
  \item \relax \textbf{Mistral} [\cite{2367571:31104403}] : A powerful 7.3 billion parameter language model, demonstrates remarkable capabilities in natural language processing tasks, outperforming larger models like Llama 2 13B across various benchmarks. Utilizing advanced techniques such as Grouped-query attention (GQA) and Sliding Window Attention (SWA), Mistral 7B achieves faster inference and handles longer sequences more efficiently. These features make it particularly suitable for ER tasks in legal documents, which often involve processing extensive texts with complex domain-specific language. The model's ability to be fine-tuned easily for specific tasks further enhances its applicability in the legal domain, where precision and context understanding are crucial. Given its superior performance and efficiency, Mistral 7B is well-equipped to improve the accuracy and effectiveness of ER systems in legal document analysis.
\end{itemize}

\section{Methodology}
In this study, we employ few-shot prompt engineering to leverage the capabilities of large language models  for judicial ER in legal documents. This technique involves crafting a single, carefully designed prompt that instructs the LLM to generate responses in a specified JSON format. The JSON response includes both the extracted text and the corresponding entity labels from the input document. This approach is particularly advantageous as it eliminates the necessity for extensive task-specific training. By directly utilizing the pre-trained LLM's advanced natural language understanding, we can efficiently identify and label entities within legal texts, streamlining the process and reducing the overhead typically associated with model training and fine-tuning.

\section{Results and Discussions}

\subsection{Experimental Setup}We evaluate the model on the InLegalNER dataset [\cite{2367571:31105196}] to rigorously assess the performance of Large Language Models on domain-specific  Entity Recognition  tasks. The InLegalNER dataset is specifically designed to encompass a comprehensive range of entities pertinent to the legal domain, thereby providing a robust benchmark for evaluating the capability of LLMs in recognizing and categorizing legal entities accurately. Table~\ref{tw-9a71c235d76b} presents a detailed breakdown of the various entities included in the dataset, offering insights into the diversity and complexity of the entity types that the models are required to identify. This evaluation aims to highlight the effectiveness of LLMs in handling the specialized terminology and context inherent in legal documents, thereby contributing to the advancement of ER methodologies in this critical domain.

\begin{table*}[!htbp]
\caption{{InLegalNER Dataset Entity Information} }
\label{tw-9a71c235d76b}
\def\arraystretch{1}
\ignorespaces 
\centering 
\begin{tabulary}{\linewidth}{p{\dimexpr.269\linewidth-2\tabcolsep}p{\dimexpr.48686666666666675\linewidth-2\tabcolsep}p{\dimexpr.2441333333333333\linewidth-2\tabcolsep}}
\tbltoprule \cellcolor[HTML]{CCCCCC}{\textbf{Named Entity}} & \cellcolor[HTML]{CCCCCC}{\textbf{Description}} & \cAlignHack \cellcolor[HTML]{CCCCCC}{\textbf{\% Occurence}}\\
\tblmidrule 
\textbf{COURT} &
  Name of any court mentioned if extracted &
  \cAlignHack 7.90\%\\
\textbf{PETITIONER} &
  Name of the petitioners / appellants / revisionist from current case &
  \cAlignHack 10.24\%\\
\textbf{RESPONDENT} &
  Name of the respondents / defendants / opposition from current case &
  \cAlignHack 12.89\%\\
\textbf{JUDGE} &
  Name of the judges &
  \cAlignHack 7.76\%\\
\textbf{LAWYER} &
  Name of the lawyers from both the parties &
  \cAlignHack 11.70\%\\
\textbf{DATE} &
  Any date mentioned in the judgment &
  \cAlignHack 6.29\%\\
\textbf{ORG} &
  Name of organizations mentioned in text apart from the court. &
  \cAlignHack 4.81\%\\
\textbf{GPE} &
  Geopolitical locations which include names of states, cities, villages &
  \cAlignHack 4.67\%\\
\textbf{STATUTE} &
  Name of the act or law mentioned in the judgement &
  \cAlignHack 6.02\%\\
\textbf{PROVISION} &
  Sections, sub-sections, articles, orders, rules under a statute &
  \cAlignHack 7.96\%\\
\textbf{PRECEDENT} &
  All the past court cases referred to in the judgement as precedent. &
  \cAlignHack 4.51\%\\
\textbf{CASE\_NUMBER} &
  All the other case numbers mentioned in the judgment (apart from precedent) &
  \cAlignHack 3.47\%\\
\textbf{WITNESS} &
  Name of witnesses in current judgment &
  \cAlignHack 2.94\%\\
\textbf{OTHER\_PERSON} &
  Name of all the persons that are not included in petitioner, respondent, judge and witness &
  \cAlignHack 8.85\%\\
\tblbottomrule 
\end{tabulary}\par 
\end{table*}

\subsection{Model Evaluation}In our study, we evaluated the performance of several state-of-the-art large language models  for the  Entity Recognition  task within legal documents. The models evaluated include LLaMA 3, Gemma, Mistral, and Phi 3. We utilized precision, recall, and F1 score as our evaluation metrics, which provide a comprehensive assessment of each model's accuracy and effectiveness in identifying and labeling entities.

\begin{table}[!htbp]
\caption{{Evaluation of the LLM Models} }
\label{tw-d6ea8cdd6343}
\def\arraystretch{1}
\ignorespaces 
\centering 
\begin{tabulary}{\linewidth}{p{\dimexpr.25\linewidth-2\tabcolsep}p{\dimexpr.2438\linewidth-2\tabcolsep}p{\dimexpr.2562\linewidth-2\tabcolsep}p{\dimexpr.25\linewidth-2\tabcolsep}}
\tbltoprule \cAlignHack \cellcolor[HTML]{CCCCCC}{Model} & \cAlignHack \cellcolor[HTML]{CCCCCC}{Precision} & \cAlignHack \cellcolor[HTML]{CCCCCC}{Recall} & \cAlignHack \cellcolor[HTML]{CCCCCC}{F1 Score}\\
\tblmidrule 
\cAlignHack LLaMA 3 &
  \cAlignHack \textbf{0.7366} &
  \cAlignHack 0.6286 &
  \cAlignHack 0.5917\\
\cAlignHack Gemma &
  \cAlignHack 0.7131 &
  \cAlignHack 0.6534 &
  \cAlignHack 0.6353\\
\cAlignHack Mistral &
  \cAlignHack 0.7097 &
  \cAlignHack \textbf{0.6628} &
  \cAlignHack \textbf{0.6376}\\
\cAlignHack Phi 3 &
  \cAlignHack 0.5975 &
  \cAlignHack 0.5617 &
  \cAlignHack 0.5440\\
\tblbottomrule 
\end{tabulary}\par 
\end{table}

\subsubsection{\textbf{LLaMA 3 Evaluation} \mbox{}\protect\newline }The LLaMA model demonstrated a precision of 0.7366, a recall of 0.6286, and an F1 score of 0.5917. While the model shows high precision, indicating a strong ability to correctly identify relevant entities, its recall is relatively lower, suggesting some missed entities within the text.

\subsubsection{Gemma Evaluation} \mbox{}\protect\newline The Gemma model yielded a precision of 0.7131, a recall of 0.6534, and an F1 score of 0.6353. Gemma's balanced precision and recall indicate a more consistent performance in identifying entities correctly and ensuring fewer missed entities, resulting in a higher F1 score compared to LLaMA.

\subsubsection{Mistral Evaluation} \mbox{}\protect\newline The Mistral model achieved a precision of 0.7097, a recall of 0.6628, and an F1 score of 0.6376. Mistral's performance is similar to Gemma, with a slightly lower precision but higher recall, which translates to a marginally better F1 score. This suggests that Mistral is effective in identifying a comprehensive set of entities while maintaining a reasonable level of accuracy.

\subsubsection{Phi 3 Evaluation} \mbox{}\protect\newline The PHI3 model showed a precision of 0.5975, a recall of 0.5617, and an F1 score of 0.5440. PHI3's lower precision and recall indicate challenges in both correctly identifying and not missing entities, resulting in the lowest F1 score among the evaluated models.

\subsection{Comparative Analysis} \mbox{}\protect\newline Overall, Mistral emerged as the best-performing model with the highest F1 score of 0.6376, closely followed by Gemma with an F1 score of 0.6353. Both models demonstrated a good balance between precision and recall, making them suitable for the NER task in legal documents. LLaMA 3, despite its higher precision, lagged in recall, indicating potential gaps in entity recognition. Phi 3 showed the least favorable performance across all metrics, suggesting it is less suited for this specific task compared to the other models evaluated.

These evaluations underscore the importance of considering both precision and recall in ER tasks, particularly in the legal domain where the accurate and comprehensive identification of entities is crucial. The results highlight Mistral and Gemma as robust options for further exploration and deployment in legal ER applications.

\section{Conclusion}
In conclusion, our study evaluated several state-of-the-art LLMs for legal entity recognition from Case Law Documents, focusing on their performance in handling domain-specific language within Indian judicial texts. Mistral and Gemma emerged as the top-performing models, showcasing balanced precision and recall crucial for accurate entity identification. These findings underscore the potential of LLMs to revolutionize ER in legal documents, offering efficient and precise entity recognition capabilities that benefit legal information management and analysis. Continued advancements in LLM architectures hold promise for further enhancing ER systems in the legal domain.
    
\section{Funding}
This study received no external funding.  
    
\section{Competing interests}
The authors declare that they have no competing interests  
    
\section{Availability of data and materials}
The used and/or during the current study (the bibliography of included studies) are available from the corresponding author upon request.  
\section*{Acknowledgements}Not applicable.

\bibliographystyle{model2-names}

\bibliography{\jobname}

\begin{thebibliography}{11}
\expandafter\ifx\csname natexlab\endcsname\relax\def\natexlab#1{#1}\fi
\expandafter\ifx\csname url\endcsname\relax
  \def\url#1{\texttt{#1}}\fi
\expandafter\ifx\csname urlprefix\endcsname\relax\def\urlprefix{URL }\fi
\providecommand{\eprint}[2][]{\url{#2}}
\providecommand{\bibinfo}[2]{#2}
\ifx\xfnm\relax \def\xfnm[#1]{\unskip,\space#1}\fi
\bibitem[{Abdin et~al.(2024)Abdin, Jacobs, Awan, Aneja, Awadallah, Awadalla,
  Bach, Bahree, Bakhtiari, Bao, Behl, Benhaim, Bilenko, Bjorck, Bubeck, Cai,
  Cai, Mendes, Chen, Chaudhary, Chen, Chen, Chen, Chen, Chopra, Dai, Giorno,
  de~Rosa, Dixon, Eldan, Fragoso, Iter, Gao, Gao, Gao, Garg, Goswami,
  Gunasekar, Haider, Hao, Hewett, Huynh, Javaheripi, Jin, Kauffmann,
  Karampatziakis, Kim, Khademi, Kurilenko, Lee, Lee, Li, Li, Liang, Liden, Liu,
  Liu, Liu, Lin, Lin, Luo, Madan, Mazzola, Mitra, Modi, Nguyen, Norick, Patra,
  Perez-Becker, Portet, Pryzant, Qin, Radmilac, Rosset, Roy, Ruwase, Saarikivi,
  Saied, Salim, Santacroce, Shah, Shang, Sharma, Shukla, Song, Tanaka, Tupini,
  Wang, Wang, Wang, Wang, Ward, Wang, Witte, Wu, Wyatt, Xiao, Xu, Xu, Xu,
  Yadav, Yang, Yang, Yang, Yang, Yu, Yuan, Zhang, Zhang, Zhang, Zhang, Zhang,
  Zhang, Zhang and Zhou}]{2367571:31104390}
\bibinfo{author}{Abdin, M.}, \bibinfo{author}{Jacobs, S.A.},
  \bibinfo{author}{Awan, A.A.}, \bibinfo{author}{Aneja, J.},
  \bibinfo{author}{Awadallah, A.}, \bibinfo{author}{Awadalla, H.},
  \bibinfo{author}{Bach, N.}, \bibinfo{author}{Bahree, A.},
  \bibinfo{author}{Bakhtiari, A.}, \bibinfo{author}{Bao, J.},
  \bibinfo{author}{Behl, H.}, \bibinfo{author}{Benhaim, A.},
  \bibinfo{author}{Bilenko, M.}, \bibinfo{author}{Bjorck, J.},
  \bibinfo{author}{Bubeck, S.}, \bibinfo{author}{Cai, Q.},
  \bibinfo{author}{Cai, M.}, \bibinfo{author}{Mendes, C.C.T.},
  \bibinfo{author}{Chen, W.}, \bibinfo{author}{Chaudhary, V.},
  \bibinfo{author}{Chen, D.}, \bibinfo{author}{Chen, D.},
  \bibinfo{author}{Chen, Y.C.}, \bibinfo{author}{Chen, Y.L.},
  \bibinfo{author}{Chopra, P.}, \bibinfo{author}{Dai, X.},
  \bibinfo{author}{Giorno, A.D.}, \bibinfo{author}{de~Rosa, G.},
  \bibinfo{author}{Dixon, M.}, \bibinfo{author}{Eldan, R.},
  \bibinfo{author}{Fragoso, V.}, \bibinfo{author}{Iter, D.},
  \bibinfo{author}{Gao, M.}, \bibinfo{author}{Gao, M.}, \bibinfo{author}{Gao,
  J.}, \bibinfo{author}{Garg, A.}, \bibinfo{author}{Goswami, A.},
  \bibinfo{author}{Gunasekar, S.}, \bibinfo{author}{Haider, E.},
  \bibinfo{author}{Hao, J.}, \bibinfo{author}{Hewett, R.J.},
  \bibinfo{author}{Huynh, J.}, \bibinfo{author}{Javaheripi, M.},
  \bibinfo{author}{Jin, X.}, \bibinfo{author}{Kauffmann, P.},
  \bibinfo{author}{Karampatziakis, N.}, \bibinfo{author}{Kim, D.},
  \bibinfo{author}{Khademi, M.}, \bibinfo{author}{Kurilenko, L.},
  \bibinfo{author}{Lee, J.R.}, \bibinfo{author}{Lee, Y.T.},
  \bibinfo{author}{Li, Y.}, \bibinfo{author}{Li, Y.}, \bibinfo{author}{Liang,
  C.}, \bibinfo{author}{Liden, L.}, \bibinfo{author}{Liu, C.},
  \bibinfo{author}{Liu, M.}, \bibinfo{author}{Liu, W.}, \bibinfo{author}{Lin,
  E.}, \bibinfo{author}{Lin, Z.}, \bibinfo{author}{Luo, C.},
  \bibinfo{author}{Madan, P.}, \bibinfo{author}{Mazzola, M.},
  \bibinfo{author}{Mitra, A.}, \bibinfo{author}{Modi, H.},
  \bibinfo{author}{Nguyen, A.}, \bibinfo{author}{Norick, B.},
  \bibinfo{author}{Patra, B.}, \bibinfo{author}{Perez-Becker, D.},
  \bibinfo{author}{Portet, T.}, \bibinfo{author}{Pryzant, R.},
  \bibinfo{author}{Qin, H.}, \bibinfo{author}{Radmilac, M.},
  \bibinfo{author}{Rosset, C.}, \bibinfo{author}{Roy, S.},
  \bibinfo{author}{Ruwase, O.}, \bibinfo{author}{Saarikivi, O.},
  \bibinfo{author}{Saied, A.}, \bibinfo{author}{Salim, A.},
  \bibinfo{author}{Santacroce, M.}, \bibinfo{author}{Shah, S.},
  \bibinfo{author}{Shang, N.}, \bibinfo{author}{Sharma, H.},
  \bibinfo{author}{Shukla, S.}, \bibinfo{author}{Song, X.},
  \bibinfo{author}{Tanaka, M.}, \bibinfo{author}{Tupini, A.},
  \bibinfo{author}{Wang, X.}, \bibinfo{author}{Wang, L.},
  \bibinfo{author}{Wang, C.}, \bibinfo{author}{Wang, Y.},
  \bibinfo{author}{Ward, R.}, \bibinfo{author}{Wang, G.},
  \bibinfo{author}{Witte, P.}, \bibinfo{author}{Wu, H.},
  \bibinfo{author}{Wyatt, M.}, \bibinfo{author}{Xiao, B.}, \bibinfo{author}{Xu,
  C.}, \bibinfo{author}{Xu, J.}, \bibinfo{author}{Xu, W.},
  \bibinfo{author}{Yadav, S.}, \bibinfo{author}{Yang, F.},
  \bibinfo{author}{Yang, J.}, \bibinfo{author}{Yang, Z.},
  \bibinfo{author}{Yang, Y.}, \bibinfo{author}{Yu, D.}, \bibinfo{author}{Yuan,
  L.}, \bibinfo{author}{Zhang, C.}, \bibinfo{author}{Zhang, C.},
  \bibinfo{author}{Zhang, J.}, \bibinfo{author}{Zhang, L.L.},
  \bibinfo{author}{Zhang, Y.}, \bibinfo{author}{Zhang, Y.},
  \bibinfo{author}{Zhang, Y.}, \bibinfo{author}{Zhou, X.},
  \bibinfo{year}{2024}.
\newblock \bibinfo{title}{{Phi-3 Technical Report: A Highly Capable Language
  Model Locally on Your Phone}}.
\bibitem[{AI@Meta(2024)}]{2367571:31103653}
\bibinfo{author}{AI@Meta}, \bibinfo{year}{2024}.
\newblock \bibinfo{title}{{Llama 3 Model Card}} .
\bibitem[{Chalkidis et~al.(2020)Chalkidis, Fergadiotis, Malakasiotis, Aletras
  and Androutsopoulos}]{2367571:31103100}
\bibinfo{author}{Chalkidis, I.}, \bibinfo{author}{Fergadiotis, M.},
  \bibinfo{author}{Malakasiotis, P.}, \bibinfo{author}{Aletras, N.},
  \bibinfo{author}{Androutsopoulos, I.}, \bibinfo{year}{2020}.
\newblock \bibinfo{title}{{LEGAL-BERT: The Muppets straight out of Law
  School}}, in: \bibinfo{editor}{Cohn, T.}, \bibinfo{editor}{He, Y.},
  \bibinfo{editor}{Liu, Y.} (Eds.), \bibinfo{booktitle}{{Findings of the
  Association for Computational Linguistics: EMNLP 2020}},
  \bibinfo{publisher}{Association for Computational Linguistics},
  \bibinfo{address}{Online}. pp. \bibinfo{pages}{2898--2904}.
\bibitem[{Jiang et~al.(2023)Jiang, Sablayrolles, Mensch, Bamford, Chaplot,
  de~las Casas, Bressand, Lengyel, Lample, Saulnier, Lavaud, Lachaux, Stock,
  Scao, Lavril, Wang, Lacroix and Sayed}]{2367571:31104403}
\bibinfo{author}{Jiang, A.Q.}, \bibinfo{author}{Sablayrolles, A.},
  \bibinfo{author}{Mensch, A.}, \bibinfo{author}{Bamford, C.},
  \bibinfo{author}{Chaplot, D.S.}, \bibinfo{author}{de~las Casas, D.},
  \bibinfo{author}{Bressand, F.}, \bibinfo{author}{Lengyel, G.},
  \bibinfo{author}{Lample, G.}, \bibinfo{author}{Saulnier, L.},
  \bibinfo{author}{Lavaud, L.R.}, \bibinfo{author}{Lachaux, M.A.},
  \bibinfo{author}{Stock, P.}, \bibinfo{author}{Scao, T.L.},
  \bibinfo{author}{Lavril, T.}, \bibinfo{author}{Wang, T.},
  \bibinfo{author}{Lacroix, T.}, \bibinfo{author}{Sayed, W.E.},
  \bibinfo{year}{2023}.
\newblock \bibinfo{title}{{Mistral 7B}}.
\bibitem[{Kalamkar et~al.(2022)Kalamkar, Agarwal, Tiwari, Gupta, Karn and
  Raghavan}]{2367571:31105196}
\bibinfo{author}{Kalamkar, P.}, \bibinfo{author}{Agarwal, A.},
  \bibinfo{author}{Tiwari, A.}, \bibinfo{author}{Gupta, S.},
  \bibinfo{author}{Karn, S.}, \bibinfo{author}{Raghavan, V.},
  \bibinfo{year}{2022}.
\newblock \bibinfo{title}{{Named Entity Recognition in Indian court
  judgments}}, in: \bibinfo{booktitle}{{Proceedings of the Natural Legal
  Language Processing Workshop 2022}}, \bibinfo{publisher}{Association for
  Computational Linguistics}, \bibinfo{address}{Abu Dhabi, United Arab Emirates
  (Hybrid)}. pp. \bibinfo{pages}{184--193}.
\bibitem[{Paul et~al.(2023)Paul, Mandal, Goyal and Ghosh}]{2367571:31103112}
\bibinfo{author}{Paul, S.}, \bibinfo{author}{Mandal, A.},
  \bibinfo{author}{Goyal, P.}, \bibinfo{author}{Ghosh, S.},
  \bibinfo{year}{2023}.
\newblock \bibinfo{title}{{Pre-trained Language Models for the Legal Domain: A
  Case Study on Indian Law}}, in: \bibinfo{booktitle}{{Proceedings of the
  Nineteenth International Conference on Artificial Intelligence and Law}},
  \bibinfo{publisher}{Association for Computing Machinery},
  \bibinfo{address}{New York, NY, USA}. pp. \bibinfo{pages}{187--196}.
\bibitem[{Team et~al.(2024)Team, Mesnard, Hardin, Dadashi, Bhupatiraju, Pathak,
  Sifre, Rivi{\`e}re, Kale, Love, Tafti, Hussenot, Sessa, Chowdhery, Roberts,
  Barua, Botev, Castro-Ros, Slone, H{\'e}liou, Tacchetti, Bulanova, Paterson,
  Tsai, Shahriari, Lan, Choquette-Choo, Crepy, Cer, Ippolito, Reid,
  Buchatskaya, Ni, Noland, Yan, Tucker, Muraru, Rozhdestvenskiy, Michalewski,
  Tenney, Grishchenko, Austin, Keeling, Labanowski, Lespiau, Stanway, Brennan,
  Chen, Ferret, Chiu, Mao-Jones, Lee, Yu, Millican, Sjoesund, Lee, Dixon, Reid,
  Miku{\l}a, Wirth, Sharman, Chinaev, Thain, Bachem, Chang, Wahltinez, Bailey,
  Michel, Yotov, Chaabouni, Comanescu, Jana, Anil, McIlroy, Liu, Mullins,
  Smith, Borgeaud, Girgin, Douglas, Pandya, Shakeri, De, Klimenko, Hennigan,
  Feinberg, Stokowiec, hui Chen, Ahmed, Gong, Warkentin, Peran, Giang, Farabet,
  Vinyals, Dean, Kavukcuoglu, Hassabis, Ghahramani, Eck, Barral, Pereira,
  Collins, Joulin, Fiedel, Senter, Andreev and Kenealy}]{2367571:31104389}
\bibinfo{author}{Team, G.}, \bibinfo{author}{Mesnard, T.},
  \bibinfo{author}{Hardin, C.}, \bibinfo{author}{Dadashi, R.},
  \bibinfo{author}{Bhupatiraju, S.}, \bibinfo{author}{Pathak, S.},
  \bibinfo{author}{Sifre, L.}, \bibinfo{author}{Rivi{\`e}re, M.},
  \bibinfo{author}{Kale, M.S.}, \bibinfo{author}{Love, J.},
  \bibinfo{author}{Tafti, P.}, \bibinfo{author}{Hussenot, L.},
  \bibinfo{author}{Sessa, P.G.}, \bibinfo{author}{Chowdhery, A.},
  \bibinfo{author}{Roberts, A.}, \bibinfo{author}{Barua, A.},
  \bibinfo{author}{Botev, A.}, \bibinfo{author}{Castro-Ros, A.},
  \bibinfo{author}{Slone, A.}, \bibinfo{author}{H{\'e}liou, A.},
  \bibinfo{author}{Tacchetti, A.}, \bibinfo{author}{Bulanova, A.},
  \bibinfo{author}{Paterson, A.}, \bibinfo{author}{Tsai, B.},
  \bibinfo{author}{Shahriari, B.}, \bibinfo{author}{Lan, C.L.},
  \bibinfo{author}{Choquette-Choo, C.A.}, \bibinfo{author}{Crepy, C.},
  \bibinfo{author}{Cer, D.}, \bibinfo{author}{Ippolito, D.},
  \bibinfo{author}{Reid, D.}, \bibinfo{author}{Buchatskaya, E.},
  \bibinfo{author}{Ni, E.}, \bibinfo{author}{Noland, E.}, \bibinfo{author}{Yan,
  G.}, \bibinfo{author}{Tucker, G.}, \bibinfo{author}{Muraru, G.C.},
  \bibinfo{author}{Rozhdestvenskiy, G.}, \bibinfo{author}{Michalewski, H.},
  \bibinfo{author}{Tenney, I.}, \bibinfo{author}{Grishchenko, I.},
  \bibinfo{author}{Austin, J.}, \bibinfo{author}{Keeling, J.},
  \bibinfo{author}{Labanowski, J.}, \bibinfo{author}{Lespiau, J.B.},
  \bibinfo{author}{Stanway, J.}, \bibinfo{author}{Brennan, J.},
  \bibinfo{author}{Chen, J.}, \bibinfo{author}{Ferret, J.},
  \bibinfo{author}{Chiu, J.}, \bibinfo{author}{Mao-Jones, J.},
  \bibinfo{author}{Lee, K.}, \bibinfo{author}{Yu, K.},
  \bibinfo{author}{Millican, K.}, \bibinfo{author}{Sjoesund, L.L.},
  \bibinfo{author}{Lee, L.}, \bibinfo{author}{Dixon, L.},
  \bibinfo{author}{Reid, M.}, \bibinfo{author}{Miku{\l}a, M.},
  \bibinfo{author}{Wirth, M.}, \bibinfo{author}{Sharman, M.},
  \bibinfo{author}{Chinaev, N.}, \bibinfo{author}{Thain, N.},
  \bibinfo{author}{Bachem, O.}, \bibinfo{author}{Chang, O.},
  \bibinfo{author}{Wahltinez, O.}, \bibinfo{author}{Bailey, P.},
  \bibinfo{author}{Michel, P.}, \bibinfo{author}{Yotov, P.},
  \bibinfo{author}{Chaabouni, R.}, \bibinfo{author}{Comanescu, R.},
  \bibinfo{author}{Jana, R.}, \bibinfo{author}{Anil, R.},
  \bibinfo{author}{McIlroy, R.}, \bibinfo{author}{Liu, R.},
  \bibinfo{author}{Mullins, R.}, \bibinfo{author}{Smith, S.L.},
  \bibinfo{author}{Borgeaud, S.}, \bibinfo{author}{Girgin, S.},
  \bibinfo{author}{Douglas, S.}, \bibinfo{author}{Pandya, S.},
  \bibinfo{author}{Shakeri, S.}, \bibinfo{author}{De, S.},
  \bibinfo{author}{Klimenko, T.}, \bibinfo{author}{Hennigan, T.},
  \bibinfo{author}{Feinberg, V.}, \bibinfo{author}{Stokowiec, W.},
  \bibinfo{author}{hui Chen, Y.}, \bibinfo{author}{Ahmed, Z.},
  \bibinfo{author}{Gong, Z.}, \bibinfo{author}{Warkentin, T.},
  \bibinfo{author}{Peran, L.}, \bibinfo{author}{Giang, M.},
  \bibinfo{author}{Farabet, C.}, \bibinfo{author}{Vinyals, O.},
  \bibinfo{author}{Dean, J.}, \bibinfo{author}{Kavukcuoglu, K.},
  \bibinfo{author}{Hassabis, D.}, \bibinfo{author}{Ghahramani, Z.},
  \bibinfo{author}{Eck, D.}, \bibinfo{author}{Barral, J.},
  \bibinfo{author}{Pereira, F.}, \bibinfo{author}{Collins, E.},
  \bibinfo{author}{Joulin, A.}, \bibinfo{author}{Fiedel, N.},
  \bibinfo{author}{Senter, E.}, \bibinfo{author}{Andreev, A.},
  \bibinfo{author}{Kenealy, K.}, \bibinfo{year}{2024}.
\newblock \bibinfo{title}{{Gemma: Open Models Based on Gemini Research and
  Technology}}.
\bibitem[{Thomas(2024)}]{2367571:31134625}
\bibinfo{author}{Thomas, A.}, \bibinfo{year}{2024}.
\newblock \bibinfo{title}{{Exploring the Power of AI-Driven Decision Making in
  the Judicial Domain: Case Studies, Benefits, Challenges, and Solutions}}.
\bibitem[{Thomas and Sangeetha(2019)}]{2367571:31134531}
\bibinfo{author}{Thomas, A.}, \bibinfo{author}{Sangeetha, S.},
  \bibinfo{year}{2019}.
\newblock \bibinfo{title}{{An innovative hybrid approach for extracting named
  entities from unstructured text data}}.
\newblock \bibinfo{journal}{{Computational Intelligence}} \bibinfo{volume}{35},
  \bibinfo{pages}{799--826}.
\bibitem[{Thomas and Sangeetha(2022)}]{2367571:31134537}
\bibinfo{author}{Thomas, A.}, \bibinfo{author}{Sangeetha, S.},
  \bibinfo{year}{2022}.
\newblock \bibinfo{title}{{Knowledge graph based question-answering system for
  effective case law analysis}}, in: \bibinfo{booktitle}{{Evolution in
  Computational Intelligence: Proceedings of the 9th International Conference
  on Frontiers in Intelligent Computing: Theory and Applications (FICTA
  2021)}}, pp. \bibinfo{pages}{291--300}.
\bibitem[{Thomas and Sivanesan(2022)}]{2367571:31134536}
\bibinfo{author}{Thomas, A.}, \bibinfo{author}{Sivanesan, S.},
  \bibinfo{year}{2022}.
\newblock \bibinfo{title}{{An adaptable, high-performance relation extraction
  system for complex sentences}}.
\newblock \bibinfo{journal}{{Knowledge-Based Systems}} \bibinfo{volume}{251},
  \bibinfo{pages}{108956}.

\end{thebibliography}

\end{document}